\documentclass[lettersize,journal]{IEEEtran}
\usepackage{amsmath,amsfonts}
\usepackage{array}
\usepackage[caption=false,font=normalsize,labelfont=sf,textfont=sf]{subfig}
\usepackage{stfloats}
\usepackage{graphicx}
\usepackage[utf8]{inputenc} %
\usepackage[T1]{fontenc}    %
\usepackage[pagebackref,breaklinks,colorlinks]{hyperref}       %
\usepackage{url}            %
\usepackage{booktabs}       %
\usepackage{amsfonts}       %
\usepackage{nicefrac}       %
\usepackage{microtype}      %
\usepackage{xcolor}         %
\usepackage[numbers,sort&compress]{natbib}

\usepackage{amsmath}
\usepackage{amssymb}
\usepackage{mathtools}
\usepackage{comment}  %
\usepackage{siunitx}
\usepackage{physics}
\usepackage{multirow}
\usepackage{mathtools}
\usepackage{utfsym}  %
\usepackage{algpseudocode}
\usepackage{algorithm}
\usepackage{amsthm}
\usepackage[capitalize]{cleveref}

\theoremstyle{definition}

\theoremstyle{remark}

\newcommand{\RR}{\mathbb{R}}

\newcommand{\eg}{e.g.}
\newcommand{\ie}{i.e.}

\begin{document}

\title{Defending against Patch-Based and Texture-Based Adversarial Attacks with Spectral Decomposition}

\author{Wei Zhang, Xinyu Chang, Xiao Li, Yiming Zhu, and Xiaolin Hu, \IEEEmembership{Senior Member, IEEE}
\thanks{This work was supported by the National Natural Science Foundation of China (No. U2341228 and No. 62576187). \textit{(Corresponding author: Xiaolin Hu.)}}%
\thanks{Wei Zhang and Xiao Li are with Department of Computer Science and Technology, Tsinghua University, Beijing 100084, China. E-mail: \{zhangw23, lixiao20\}@mails.tsinghua.edu.cn.}%
\thanks{Xinyu Chang is with School of Software, Tsinghua University, Beijing 100084, China. E-mail: chang-xy25@mails.tsinghua.edu.cn.}%
\thanks{Yiming Zhu is with School of Computer and Communication Engineering, University of Science and Technology Beijing, Beijing 100083, China. E-mail: u202241171@xs.ustb.edu.cn.}%
\thanks{Xiaolin Hu is with Department of Computer Science and Technology, Institute for AI, BNRist, IDG/McGovern Institute for Brain Research, Tsinghua University, Beijing 100084, China, also with Chinese Institute for Brain Research (CIBR), Beijing 100010, China. E-mail: xlhu@tsinghua.edu.cn.}}

\markboth{}%
{Zhang \MakeLowercase{\textit{et al.}}: Defending against Patch-Based and Texture-Based Adversarial Attacks with Spectral Decomposition}

\IEEEpubid{\copyright~2026 IEEE}

\maketitle

\begin{abstract}
    Adversarial examples present significant challenges to the security of Deep Neural Network (DNN) applications.
    Specifically, there are patch-based and texture-based attacks that are usually used to craft physical-world adversarial examples, posing real threats to security-critical applications such as person detection in surveillance and autonomous systems, because those attacks are physically realizable.
    Existing defense mechanisms face challenges in the adaptive attack setting, i.e., the attacks are specifically designed against them.
    In this paper, we propose Adversarial Spectrum Defense (ASD), a defense mechanism that leverages spectral decomposition via Discrete Wavelet Transform (DWT) to analyze adversarial patterns across multiple frequency scales. The multi-resolution and localization capability of DWT enables ASD to capture both high-frequency (fine-grained) and low-frequency (spatially pervasive) perturbations.
    By integrating this spectral analysis with the off-the-shelf Adversarial Training (AT) model, ASD provides a comprehensive defense strategy against both patch-based and texture-based adversarial attacks.
    Extensive experiments demonstrate that ASD+AT achieved state-of-the-art (SOTA) performance against various attacks, outperforming the APs of previous defense methods by 21.73\%, in the face of strong adaptive adversaries specifically designed against ASD. Code available at: \url{https://github.com/weiz0823/adv-spectral-defense}.
\end{abstract}


\section{Introduction}
\label{sec:intro}
\IEEEPARstart{D}{espite} the remarkable generalization power of DNNs in various vision tasks, DNNs are known to be vulnerable to adversarial examples \citep{goodfellow2014fgsm, madry2017pgd, adversarial-patch, kurakin2018phys-adv, thys2019fooling, hu2023camou}.
Conventional adversarial attacks consider adding human-imperceptible adversarial noises that are bounded by some $l_p$ norm \citep{goodfellow2014fgsm, madry2017pgd, carlini2017cw}, called digital adversarial attacks.
Meanwhile, there are patch-based and texture-based attacks \citep{kurakin2018phys-adv, thys2019fooling, xu2020adv-tshirt, wu2020adv-cloak, hu2021nat-patch, hu2022adv-texture, hu2023camou} which define perturbation bounds different from any $l_p$ norm bound.
These attacks constrain the adversarial patterns to some predetermined area,
and craft localized adversarial patterns which are often perceptible to humans.
Specifically, patch-based attacks craft localized adversarial patterns within a fixed region (\eg, a square patch).
Texture-based attacks craft more pervasive adversarial perturbations that spread across the entire surface of the object, \eg, adversarial modifications to clothing textures that cover the entire garment.
Both patch-based and texture-based attacks can be implemented in the physical world and are thereby called physically realizable attacks.
The physically realizable natures of these attacks make them pose real threats to security-critical applications such as person detection in surveillance and autonomous systems, where an adversary can either hold a patch or wear adversarial clothes to fool the system.

\begin{figure}[!t]
    \centering
    \includegraphics[width=0.9\linewidth]{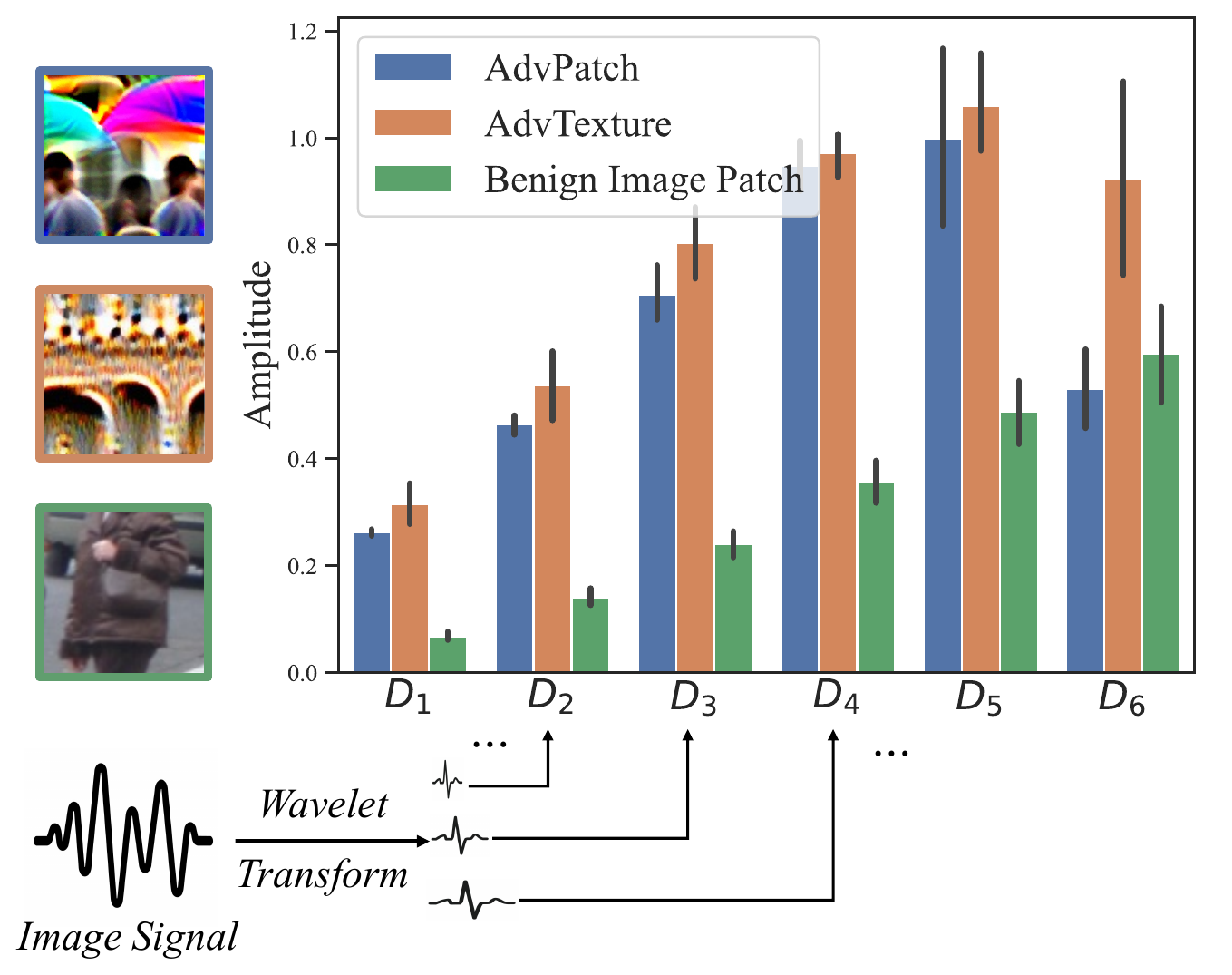}
    \caption{Spectral analysis of different physically realizable adversarial attacks with Discrete Wavelet Transform (DWT). $D_1$ to $D_6$ represent the detail (high-frequency) sub-bands obtained from multi-level DWT, with $D_1$ being the highest frequency. The amplitudes correspond to absolute values of the wavelet coefficient components at each decomposition level. They capture image edges, textures, and local spatial variations at different scales. The figure highlights abnormal spectral amplitudes caused by physical adversarial perturbations. Error bars: \SI{95}{\percent} confidence intervals.
    Upper left: An adversarial patch (AdvPatch), an adversarial texture (AdvTexture, cropped), and a benign image patch, visualized. Bottom: A sketch of DWT.}
    \label{fig:fsd_motivation}
\end{figure}

To mitigate these threats, several defense methods \citep{lgs-naseer2019,ape-kim2022,sac-liu2022,wu2024napguard,jedi-tarchoun2023} against physically realizable adversarial examples have been proposed.
However, existing defenses typically suffer from two key limitations:
(1) They face challenges in the adaptive attack setting, where the attack strategies are specifically designed to bypass the defense \citep{obfuscated,tramer2020adaptive}.
(2) They primarily target traditional patch-based attacks \citep{thys2019fooling, xu2020adv-tshirt}.
\IEEEpubidadjcol
When evaluated against more physically realizable attack forms such as texture-based attacks \citep{hu2022adv-texture, hu2023camou}, these defense methods are less effective (see \cref{sec:result_main}).
For example, NAPGuard \citep{wu2024napguard} constructed an adversarial patch detection dataset and trained detectors to detect the patch. While effective against non-adaptive patches, the lack of evaluation in the adaptive attack setting made it less effective when the patches were adaptively crafted to evade the patch detector. Moreover, texture-based adversarial patterns are more pervasive, making the patch detector harder to localize them.
Therefore, there is a pressing need for a defense strategy that can account for a range of adversarial characteristics and is effective under adaptive attack.

We notice that both patch-based and texture-based attacks are localized perturbations, and the perturbation patterns differ from normal patterns in benign image areas.
This prompts us to analyze the local patterns to understand their attack mechanisms, and DWT \citep{dwt} is an appropriate tool, which decomposes the adversarial patterns into different frequency components.
DWT offers a key advantage over the commonly used Discrete Fourier Transform (DFT): it captures localized frequency information, making it particularly effective for analyzing localized adversarial patterns.
With this tool, we can investigate the spectral characteristics of physically realizable adversarial attacks.

\Cref{fig:fsd_motivation} shows the DWT amplitudes across various frequency components of the adversarial examples crafted by a typical patch-based attack, AdvPatch \citep{thys2019fooling}, and a typical texture-based attack, AdvTexture \citep{hu2022adv-texture}, along with the frequency components of benign image patches.
It is seen that both adversarial patterns exhibit significantly larger amplitudes, suggesting that adversarial perturbations are identifiable in the frequency domain by their unique frequency signatures. This spectral distinction highlights a potential defense strategy: \emph{by localizing these high-amplitude regions, we may find a method to mitigate the impact of these adversarial attacks}.

Based on the analysis, we propose ASD, a novel defense mechanism designed to protect object detectors against patch-based and texture-based adversarial attacks.
ASD has a masking module, called ASD Masking, to detect and mask out potential adversarial regions. Then the preprocessed images are fed into an off-the-shelf AT detector to offer robustness against low-amplitude adversarial patterns. ASD level aggregation is finally used to adapt to different object sizes in the images.
Specifically, ASD uses the tool of DWT to decompose the input image into various frequency components and incorporates all of them to detect and localize adversarial patterns of the attacks.
The multi-resolution and localization capability of the spectral decomposition enables ASD to capture both high-frequency (fine-grained) and low-frequency (spatially pervasive) perturbations, providing a defense method that has the potential to defend against both patch-based and texture-based attacks. Furthermore, by integrating this spectral analysis with an off-the-shelf AT model that provides robustness against $l_\infty$-norm bounded attacks, ASD provides \emph{a comprehensive defense strategy against both high-amplitude adversarial perturbations and low-amplitude ones}, making it effective in the adaptive attack setting.

Extensive experiments demonstrate that ASD+AT effectively defends against both patch-based and texture-based attacks. Against strong adaptive adversaries, ASD achieves SOTA performance with over \SI{21.73}{\percent} improvement on AP\textsubscript{50} over existing defense methods.

Our main contributions can be summarized as follows:
\begin{itemize}
    \item We propose a novel defense strategy based on spectral decomposition and multi-frequency integration to defend against both patch-based and texture-based adversarial attacks.
    \item The proposed method, ASD, is compatible with the off-the-shelf AT model, enabling a comprehensive defense strategy that addresses patch-based and texture-based adversarial attacks in the adaptive attack setting, combining the strengths of both spectral analysis and adversarial training.
    \item Experimental results demonstrate that ASD+AT significantly outperforms existing defense methods against both patch-based and texture-based attacks.
\end{itemize}

\section{Related Work}
\label{sec:related}

\subsection{Adversarial Attacks}

Early attempts to create adversarial examples in the digital world involved adding human-imperceptible noise to image pixels \citep{goodfellow2014fgsm,madry2017pgd,carlini2017cw,kurakin2018phys-adv}. However, manipulating pixel values requires hacking into the camera system. A more practical scenario involves manipulating only the target object, such as applying adversarial patches \citep{adversarial-patch,thys2019fooling,xu2020adv-tshirt,wu2020adv-cloak,hu2021nat-patch}. One extensively studied task in this area is the person detection task \citep{nguyen2016human-det-survey,wei2024phys-adv-survey}, which utilizes object detection models to localize people in images. The first attack in this area was AdvPatch \citep{thys2019fooling}, which involved transforming and applying a universal image patch onto each person in the images from the training dataset. A successful attack resulted in the object detector failing to localize people in the image when the adversarial patch was applied. AdvTshirt \citep{xu2020adv-tshirt} printed adversarial patches on T-shirts to render the people undetectable, while \citet{wu2020adv-cloak} similarly aimed to create an invisibility cloak.

\begin{figure}[!t]
    \centering
    \includegraphics[width=0.98\linewidth]{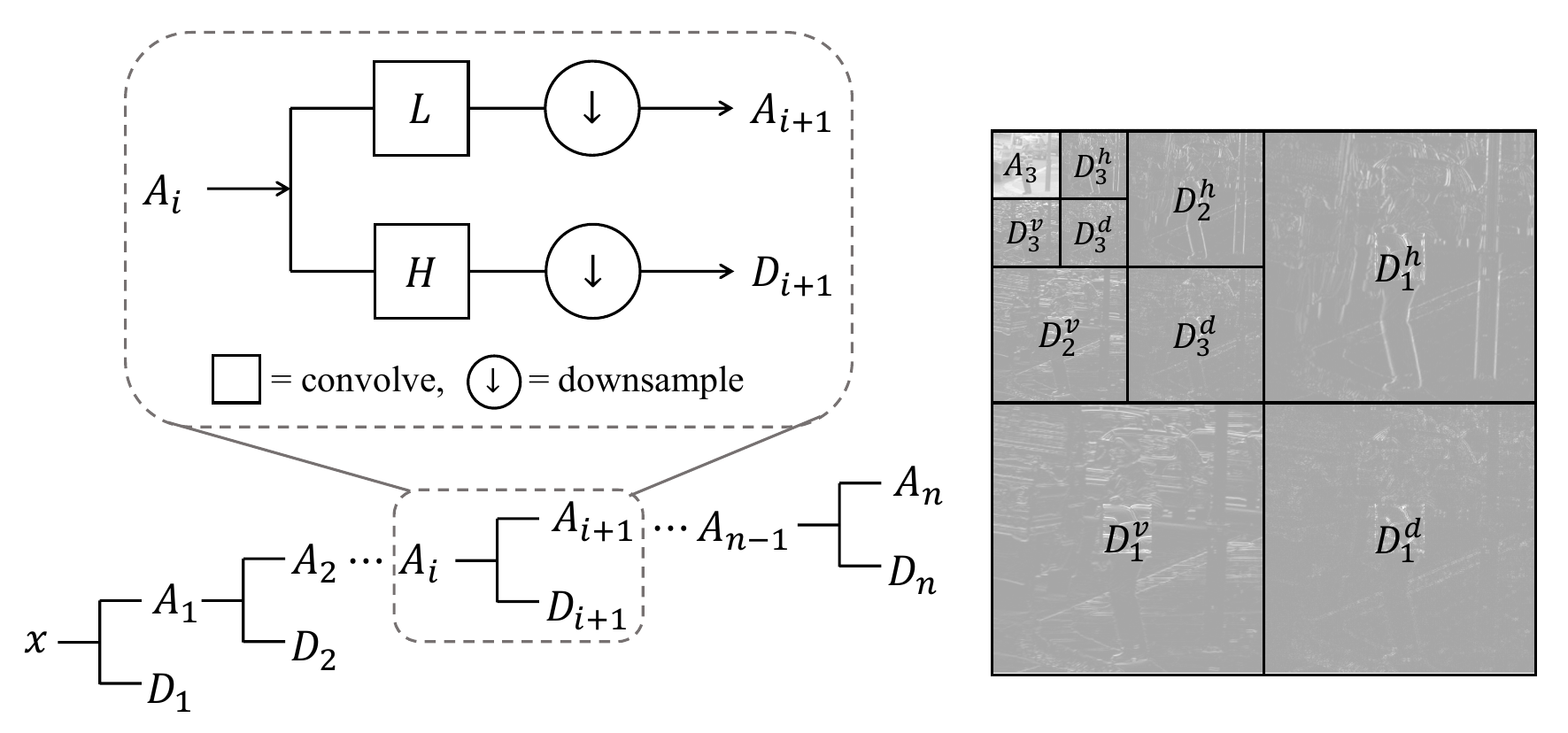}
    \caption{Illustration of the DWT decomposition. Left: The 1D hierarchical decomposition. Right: The outputs of the decomposition on a 2D image, called the 2D DWT pyramid.}
    \label{fig:illust_dwt}

\end{figure}

Recent studies \citep{hu2022adv-texture,hu2023camou} also crafted physical adversarial examples to fool person detectors in the form of adversarial textures, which covered the whole of clothes.
AdvTexture \citep{hu2022adv-texture} ensured a 360-degree adversarial effect by tiling the adversarial pattern on the whole area of the T-shirt.
AdvCaT \citep{hu2023camou} made adversarial camouflage clothes that were both adversarial and natural-looking.

Besides, there are also attack methods aiming at adding imperceptible perturbation to image to fool object detectors \citep{zhou2025numbod,li2024lgp}.
They differ from physically realizable attack form by not introducing visible features to human eyes.
Note that these attacks cannot be implemented in the physical world, since they add small perturbation to image pixels.

\subsection{Adversarial Defenses}

To defend against physically realizable adversarial attacks, various defense methods have been proposed \citep{lgs-naseer2019,adv-yolo-ji2021,ape-kim2022,sac-liu2022,jedi-tarchoun2023,bunzel2023entropy,patchzero,over-activation,wu2024napguard,jing2024pad}. To the best of our knowledge, all these methods have been evaluated solely against adversarial patch attacks. One prevailing paradigm for defending against these attacks is to localize adversarial patches and mask or inpaint those areas as a preprocessing step. For example, LGS \citep{lgs-naseer2019} proposed identifying adversarial regions based on image gradient criteria. Jedi \citep{jedi-tarchoun2023} identified high-entropy areas as potential adversarial regions. SAC \citep{sac-liu2022} trained a segmentation model to delineate patch areas, and NAPGuard \citep{wu2024napguard} developed a patch detection model for localization. Additionally, some methods \citep{ape-kim2022, over-activation} analyzed the internal feature maps of models when processing adversarial examples. 

Adversarial Training (AT) \citep{madry2017pgd} is a prevailing method to defend against digital adversarial attacks \citep{goodfellow2014fgsm,madry2017pgd,carlini2017cw,autoattack} for classification models. Recently, \citet{li2023od-at} applied AT to object detection models, achieving SOTA robustness against digital adversarial attacks with perturbations bounded by the \( l_p \) norm. We used the checkpoints provided by \citet{li2023od-at} to evaluate the performance of these AT models against physically realizable adversarial attacks.

\subsection{Adversarial Defenses Based on DWT}

There are various defense methods \citep{liu2019fd-jpeg,chen2021amplitude-phase,chen2022shapley,huang2023mae,jung2023predicting,naderi2023lpf-defense,chaudhury2024learned-in-freq-domain,agarwal2020image,sarvar2023defense,meenakshi2022adviris} that defend against adversarial attacks with frequency domain knowledge. They mainly use denoising-based pipelines and target digital or imperceptible adversarial attacks. They use time-frequency transformations such as discrete Fourier transform, discrete cosine transform, or DWT to denoise the input image. The pipelines consist of three major steps: transform into frequency domain, filter on high-frequency bands, inverse transform back to pixel domain. Moreover, all of them target the threat model of $l_p$-norm bounded attacks. To filter out adversarial noises while preserving original image semantic, the defense methods focus on purifying high-frequency band only.
Since patch-based and texture-based attacks create larger color blocks, it is unlikely for the denoising methods to get rid of them. Among them, FD-JPEG \citep{liu2019fd-jpeg} is a representative frequency-based defense method. FD-JPEG is based on JPEG compression, which is a frequency-domain compression method. FD-JPEG transforms images into the frequency domain to compress them with minimal perceptible loss in quality, then transform the distilled features back to pixel domain.

\section{Frequency Components of Physically Realizable Adversarial Examples}

\subsection{Preliminary}
\label{sec:dwt_method}

We first explain why we use DWT instead of the more commonly used DFT, then we review the formulation of 1D DWT. Finally, we review the 2D DWT applied to images.

\subsubsection{From DFT to DWT}
The DFT transforms a discrete signal into an array of numbers in the frequency domain.
The DFT captures global information of the signal and decomposes its different frequency components,
and is used by many previous works \citep{liu2019fd-jpeg,chen2022shapley,jung2023predicting,chaudhury2024learned-in-freq-domain} to develop defense methods against digital adversarial attacks.
However, physically realizable attacks are typically localized perturbations. In this context, the DWT is a more appropriate tool. The DWT employs localized wavelets as a basis to capture localized oscillations \citep{dwt}. Through downsampling in the DWT decomposition, the DWT captures higher-frequency components with finer time resolution and lower-frequency components with coarser time resolution. 

\subsubsection{1D DWT}
The DWT decomposition is formulated as follows. Starting from a 1D signal \( x \), the outputs consist of an array of approximation coefficients \( A \) and detail coefficients \( D \). The decomposition follows a hierarchical process. The approximation at level \( i \), denoted as \( A_i \), is further decomposed into the next-level detail \( D_{i+1} \) and approximation \( A_{i+1} \). A low-pass filter \( L \) is used to extract low-frequency information \( A_{i+1} \), while a high-pass filter \( H \) is used to extract high-frequency information \( D_{i+1} \). \cref{fig:illust_dwt} (left) shows the hierarchical process of DWT decomposition.

The decomposition process of DWT has two primary operations: convolution and downsampling, denoted as \( * \) and \( \mathrm{DS} \), respectively. Given the approximation \( A_i \), the updates are defined as follows:
\begin{equation}
    A_{i+1} = \mathrm{DS}(A_i * L), \qquad
    D_{i+1} = \mathrm{DS}(A_i * H).
\end{equation}
Due to the presence of the \( \mathrm{DS} \) operation, as \( i \) increases, the lengths of \( D_i \) and \( A_i \) both decrease; \( D_i \) represents lower-frequency details, while \( A_i \) becomes a coarser approximation. The decomposition starts from \( A_0 = x \). At the maximum expansion level \( n \), the final outputs of the decomposition are \( D_1, D_2, \ldots, D_n, A_n \), which correspond to details from the high-frequency component to the low-frequency component. The remaining approximation is \( A_n \). Together, these coefficients comprise the complete information of the original signal \( x \).

The frequency domain representation, which consists of the coefficients from the DWT, can be used to reconstruct the original signal via the Inverse Discrete Wavelet Transform (IDWT) process. In reconstruction, downsampling is replaced with upsampling, denoted as \( \mathrm{US} \). The inverse filters \( L^{-1} \) and \( H^{-1} \) are employed. Given \( A_i \) and \( D_i \), the previous level approximation \( A_{i-1} \) is reconstructed as:
\begin{equation}
    A_{i-1} = \mathrm{US}(A_i) * L^{-1} + \mathrm{US}(D_i) * H^{-1}.
\end{equation}
The reconstruction through IDWT starts from level \( n \) and ultimately yields \( A_0 = x \).

\subsubsection{2D DWT on images}
In this work, we use 2D DWT to decompose images. An image is viewed as a 2D signal, and DWT decomposition is done on the spatial frequency domain. The main difference between 2D DWT and 1D DWT is the number of convolutions and downsampling per level.
From approximation $A_{i}$, the decomposition is first performed along the rows, generating approximation $A^{r}_{i}$ and details $D^{r}_{i}$. Two more decompositions are performed along the columns. $A^{r}_{i}$ is decomposed into approximation $A_{i+1}$ and horizontal details $D^{h}_{i+1}$. $D^{r}_{i}$ is decomposed into vertical details $D^{v}_{i+1}$ and diagonal details $D^{d}_{i+1}$. Details at level $i+1$ consist of 3 channels, $D^{h}_{i+1}, D^{v}_{i+1}, D^{d}_{i+1}$, and approximation $A_{i+1}$ is similar to the 1D case.
By DWT decomposition, a pyramid representation of an input image is built (\cref{fig:illust_dwt}, right).

\subsubsection{Haar wavelet}
The DWT requires predetermined wavelet filters \( L \) and \( H \) to complete the transformation. In this paper, we utilize the Haar wavelet \citep{haar-wavelet}, which is commonly used in signal processing for DWT. Specifically, the Haar wavelet is characterized by a low-pass filter given by \( L = \left[ \frac{\sqrt{2}}{2}, \frac{\sqrt{2}}{2} \right] \) and a high-pass filter given by \( H = \left[ -\frac{\sqrt{2}}{2}, \frac{\sqrt{2}}{2} \right] \). The downsampling operation \( \mathrm{DS} \) for the Haar wavelet selects the odd indices, thereby reducing the length of the array by a factor of 2.

\subsection{Spectral Analysis Results of Physically Realizable Adversarial Examples}
\label{sec:motivation_freq}
Adversarial patches or textures are difficult to detect by examining the image alone, as any texture or pattern can be printed on clothing, which minimally interferes with human detection and identification of individuals. Using the DWT, we analyze whether specific patterns in frequency components can be used to enhance the robustness of detectors against physically realizable adversarial attacks. We decomposed various adversarial patches, as shown in \cref{fig:fsd_motivation}. Both AdvPatch and AdvTexture targeted the Faster R-CNN \citep{frcnn} detector. We cropped the tiled texture pattern of AdvTexture to match the size of the patch. For both attack methods, we crafted five adversarial patches using different random seeds, and the results presented reflect the averages of these five repetitions. The benign image patches correspond to 16 image patches randomly sampled from the Inria Person \citep{inria} dataset. The detail coefficients \( D_i \) are computed as the average absolute value across three directions: \( D_i^h \), \( D_i^v \), and \( D_i^d \), and are scaled by \( \frac{10}{2^i} \) for better visualization.

The results in \cref{fig:fsd_motivation} indicate that both AdvPatch and AdvTexture exhibit significantly higher DWT amplitudes than benign image patches across frequency components \( D_1 \) to \( D_5 \). For \( D_6 \), the amplitudes of AdvPatch and benign image patches are comparable. A more detailed analysis of this phenomenon will be provided in \cref{sec:reinforce_at}. These findings suggest that a spectral perspective based on DWT may be leveraged to defend against physically realizable attacks.

\section{Adversarial Spectrum Defense}
\label{sec:fsd_method}
\begin{figure*}[t]
    \centering
    \includegraphics[width=0.98\linewidth]{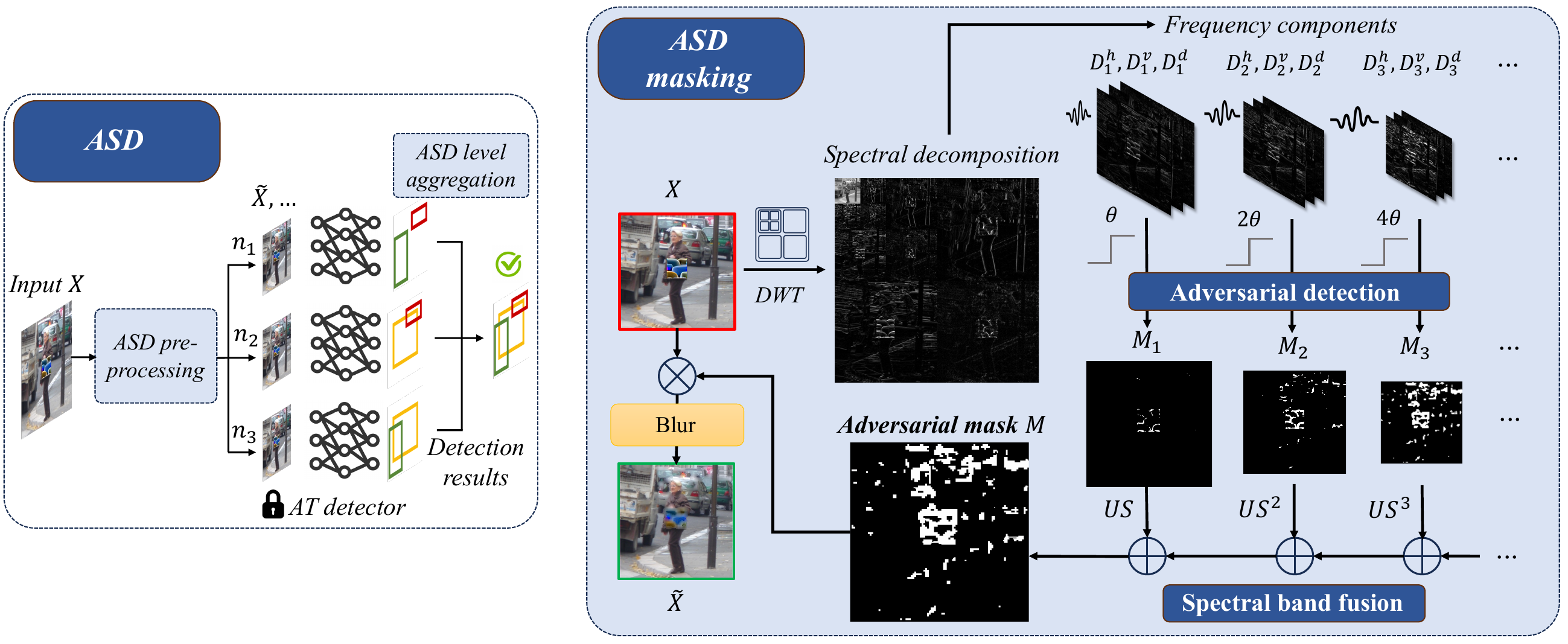}
    \caption{Overview of the pipeline of the proposed ASD method. ASD has a masking module, called \emph{ASD Masking}, to detect and mask out possible adversarial regions with $n$-level DWT decomposition. Then the preprocessed images are fed into the AT detector to offer robustness against low-amplitude adversarial patterns. By setting DWT level to $n_1, n_2, n_3, \ldots$, ASD level aggregation is finally used to adapt to different object sizes in the images. The block on the right shows the details of ASD Masking, which is based on spectral decomposition.}
    \label{fig:illust_pipeline}

\end{figure*}

In this section, we describe the pipeline of ASD. An overview of the pipeline is provided in \cref{fig:illust_pipeline}.

\subsection{Multi-Spectral Adversarial Region Masking}
\label{sec:fsd_method_single_level}
\subsubsection{Adversarial region detection in each frequency component}
Suppose that the input image $X \in \RR^{H_0\times W_0 \times C}$ has dimensions divisible by $2^n$, \ie, $2^n | H_0, 2^n | W_0$, where $n$ is the maximum level of DWT decomposition. If not, padding is applied to ensure divisibility.
We first convert the RGB image into grayscale $X^\prime \in \RR^{H_0\times W_0}$.
Then applying 2D DWT to \( X' \) yields detail coefficients at different levels, $\Bqty{D_i^h, D_i^v, D_i^d}_{i=1}^n$, where $D_i^h, D_i^v, D_i^d \in \RR^{H_i\times W_i}$. At level $i$, the sizes of the coefficients are $H_i=\flatfrac{H_0}{2^i}, W_i=\flatfrac{W_0}{2^i}$.
We stack the horizontal details $D_i^h$, the vertical details $D_i^v$, and the diagonal details $D_i^d$ into three-channel DWT feature $D_i=\mathrm{stack}(D_i^h, D_i^v, D_i^d) \in \RR^{H_i\times W_i\times 3}$. The $l_2$ norm along the channel dimension of $D_i$ is computed, producing spectral intensity map $F_i \in \RR^{H_i\times W_i}$ at level $i$.

Since adversarial patterns in the physical world usually cover continuous regions, we smooth the spectral feature intensity map $F_i$ with $F^\prime_i=\mathrm{smooth}(F_i)$ to reduce noise.
Specifically, we apply $\mathrm{smooth}$ as averaging over a $3\times 3$ neighborhood.
The adversarial region is detected from smoothed spectral feature $F^\prime_i$ with a threshold $\theta_i$. Pixels where $F^\prime_i>\theta_i$ are considered adversarial, resulting in an adversarial region mask $M_i\in\Bqty{0,1}^{H_i\times W_i}$, where these pixels are set to 1 and all other pixels are set to 0. In the Haar wavelet, the low-pass filter is $L=\bqty{\frac{\sqrt{2}}{2}, \frac{\sqrt{2}}{2}}$. Therefore, the low-frequency approximation $A_{1}$, which is the input of next-level decomposition, is amplified by a factor of 2.
For example, a constant 2D signal $f(x,y)=c\in \RR$ is decomposed into $A_{1}=2c$ and $D_1^h=D_1^v=D_1^d=0$. Therefore, the threshold $\theta_i$ at level $i$ should be set to $\theta_i=2^{i-1}\theta$, where \( \theta \) is a hyper-parameter of the ASD framework.

\subsubsection{Spectral band fusion} The adversarial region masks $M_1,M_2,\ldots,M_n$ are the adversarial detection results in different spectral bands. Without much constraint, adversarial patterns lie in different spectral bands, so it is necessary to fuse different spectral bands. With the produced mask $M_i\in\Bqty{0,1}^{H_i\times W_i}$ and sizes $H_i=\flatfrac{H_0}{2^i}, W_i=\flatfrac{W_0}{2^i}$, all masks are fused with upsampling to the original image size and union operation.
Formally, the spectral band fusion is
\begin{equation}
    M = \bigcup_{i=1}^n \mathrm{US}^i(M_i),
\end{equation}
where $\mathrm{US}^i$ denotes $i$ repeated application of the upsampling operation $\mathrm{US}$.
The final mask $M\in\Bqty{0,1}^{H_0\times W_0}$ is the detected adversarial region mask in input image $X$.

\begin{figure}[t]
    \centering
    \subfloat[]{\includegraphics[width=0.35\linewidth]{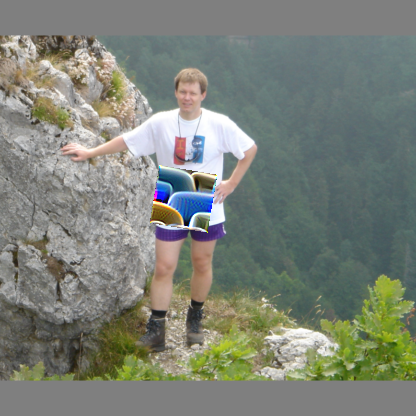}}
    \hfil
    \subfloat[]{\includegraphics[width=0.35\linewidth]{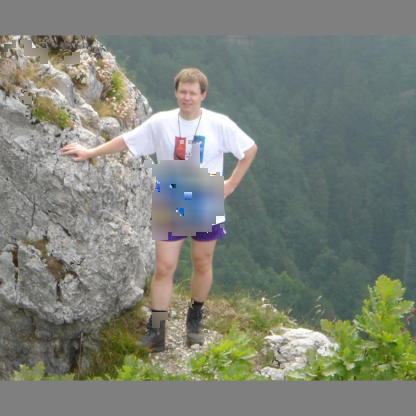}}
    \hfil
    \subfloat[]{\includegraphics[width=0.35\linewidth]{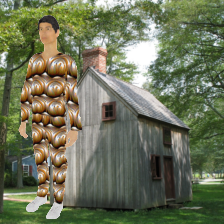}}
    \hfil
    \subfloat[]{\includegraphics[width=0.35\linewidth]{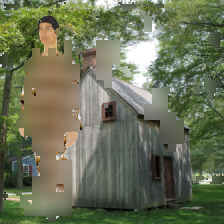}}
    \caption{Visualization of adversarial examples with patch-based or texture-based attacks, and the images processed by ASD. (a) AdvPatch \citep{thys2019fooling} example. (b) AdvPatch processed by ASD. (c) AdvTexture \citep{hu2022adv-texture} example. (d) AdvTexture processed by ASD.
    }
    \label{fig:vis_examples}

\end{figure}

With the adversarial mask, Gaussian blurring is applied on the detected adversarial regions, forming our preprocess on the input image. Example of processed images are provided in \cref{fig:vis_examples}.

\subsection{ASD Level Aggregation}
\label{sec:method_level_agg}
\begin{algorithm}[t]
	\caption{Object detection with ASD.}
    \label{alg:asd_main}
	\begin{algorithmic}[1]
	\Require Image $X$ to be detected, AT detector $f(X)=\vb{b}$, 2D DWT decomposition function $g(X; n)=D^h_1,D^v_1,D^d_1,D^h_2,D^v_2,D^d_2,\ldots,D^h_n,D^v_n,D^d_n$, a $\mathrm{grayscale}$ function that converts RGB image into grayscale, a $\mathrm{smooth}$ function, the spatial upsampling function $\mathrm{US}$, a blurring function $\mathrm{blur}$, the non-max-suppression function $\mathrm{nms}$, ASD threshold $\theta$, a series of decomposition levels $n_1, \ldots, n_k$, where $k$ is also a user-selected value.
        \Ensure Bounding boxes $\hat{\vb{b}}$.
        \State $X^\prime=\mathrm{grayscale}(X)$
        \For{$i = 1, \ldots, k$}
        \State $D^h_1,D^v_1,D^d_1,D^h_2,D^v_2,D^d_2,\ldots,D^h_{n_i},D^v_{n_i},D^d_{n_i} \gets g(X^\prime; n_i)$
        \For{$j = 1, \ldots, n_i$}
        \State $D_j \gets \mathrm{stack}(D^h_j,D^v_j,D^d_j)$
        \State $F_j \gets \norm{D_j}_2$
        \Comment $\norm{\cdot}_2$ is applied along the channel dimension.
        \State $F_j^\prime \gets \mathrm{smooth}(F_j)$
        \State $\theta_j<2^{j-1}\theta$
        \State $M_j \gets \mathrm{1}_{\Bqty{F_j^\prime>\theta_j}} $
        \EndFor
        \State $\hat{M}_i \gets \bigcup_{j=1}^{n_i} \mathrm{US}^j(M_j)$
        \Comment Spectral band fusion.
        \State $\tilde{X}_i \gets X \otimes (1-\hat{M}_i) + \mathrm{blur}(X) \otimes \hat{M}_i$
        \Comment This line completes the ASD Masking.
        \State $\vb{b}_i \gets f(\tilde{X}_i)$
        \Comment Single-level detection.
        \EndFor
        \State $\hat{\vb{b}} \gets \mathrm{nms}(\vb{b}_1, \ldots, \vb{b}_k)$
        \Comment ASD level aggregation.
	\end{algorithmic}
\end{algorithm}

In object detection tasks, the size of an object relative to the input image can vary significantly. For instance, a person positioned farther from the camera occupies fewer pixels in the image. Meanwhile, the DWT decomposition at level $i$ consistently corresponds to the same spectral band. As a result, an identical adversarial pattern may exist in different spectral bands depending on the object's distance from the camera.

Another consideration is the overlap between the frequency of the adversarial pattern and that of benign objects. The detailed textures of the object, \eg, conspicuous T-shirt patterns, are typically unimportant to the detection of the object, and false positive adversarial detection on the texture is therefore not a major concern. However, if the decomposition level $n$ is too large, the frequency of detail coefficients $D_n$ may cover the object frequency, \ie, the object size could be smaller than half wavelength. In this case, the contours of the object may be misidentified as adversarial features, potentially leading to the entire object being masked out. With the whole object masked out, the detector is unable to detect the object, and this is not the desired result. Large objects may have large adversarial patterns that lie in low-frequency components, while small objects have low base frequencies. In order to detect various sizes in the same image, results from different decomposition levels need to be aggregated.

To address these scale-dependent effects, we aggregate detection results obtained from multiple decomposition levels. Specifically, given an input image $X$, we apply the preprocessing procedure described in \cref{sec:fsd_method_single_level} for $k$ times, using different decomposition levels $n_1, n_2, \ldots, n_k$. This produces a set of preprocessed images $\tilde{X}_1, \tilde{X}_2, \ldots, \tilde{X}_k$. Each preprocessed image is independently fed into the object detector, generating a set of detection results for each level.
All detection results from the $k$ preprocessed images are merged, and a Non-Maximum Suppression (NMS) operation is applied to remove redundant bounding boxes. The remaining detections constitute the final detection results for the original input image $X$. We give the pseudo-code of the complete ASD procedure in \cref{alg:asd_main}.

\subsection{Integrating ASD with AT Model}
\label{sec:reinforce_at}
\begin{figure}[t]
    \centering
    \includegraphics[width=0.8\linewidth]{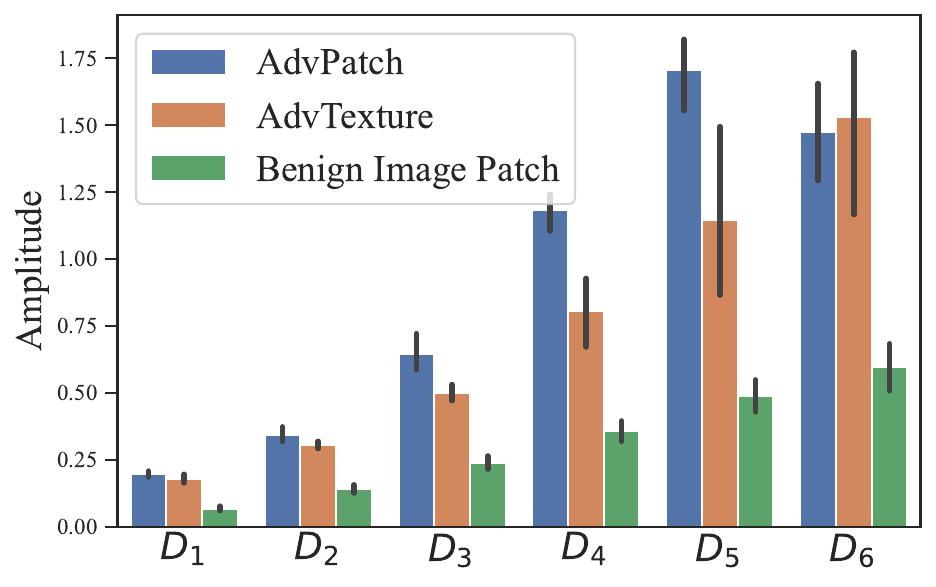}

    \caption{The DWT amplitude of different physically realizable adversarial examples against the adversarially trained Faster R-CNN. $D_1$ to $D_6$ correspond to different frequency levels, from higher frequency to lower frequency. Error bar: \SI{95}{\percent} confidence interval.}
    \label{fig:motivation_at}
\end{figure}

\begin{figure}[t]
    \centering
    \includegraphics[width=0.95\linewidth]{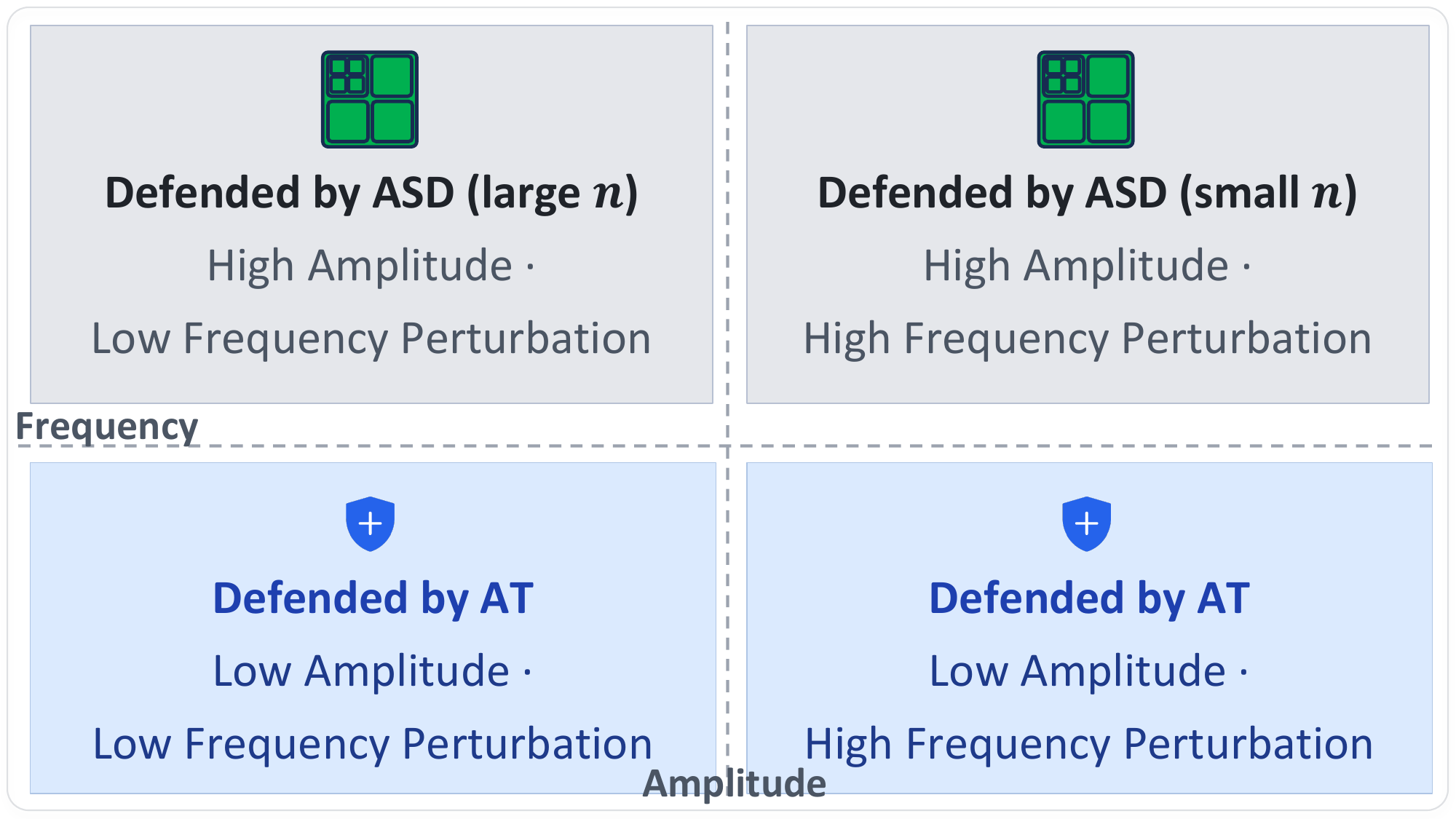}
    \caption{Frequency-amplitude division of ASD and AT.}
    \label{fig:illust_full_spectrum}
\end{figure}

In \cref{sec:motivation_freq}, we analyze different adversarial attacks and examine the different of DWT amplitude between adversarial patches and benign image patches. However, an adaptive attacker may attempt to evade the ASD masking module by designing adversarial perturbations with low DWT amplitude.

A natural class of such attacks corresponds to low-magnitude perturbations in the image space, namely $l_p$-norm bounded adversarial attacks. These attacks constrain the perturbation magnitude by $\norm{X^*-X}_p\le\epsilon$, where $X$ is the benign image, $X^*$ is the adversarial image, and $\epsilon$ is the perturbation bound. Among them, $l_\infty$-bounded attacks are widely studied, and adversarial training (AT) \citep{madry2017pgd,li2023od-at} on object detection models is a commonly used defense strategy.

Importantly, perturbations bounded in the $l_\infty$ norm also impose constraints on the DWT amplitudes. Specifically, if $\norm{X^*-X}_\infty\le\epsilon$, the DWT detail coefficients satisfy \( D_i\le 2^{i}\epsilon \) for each decomposition level $i$ (see Appendix for full proof).
This result implies that perturbations with small DWT amplitudes necessarily correspond to small pixel-space perturbations. Therefore, an attacker attempting to evade ASD by constraining the DWT amplitude is effectively restricted to the class of low-magnitude perturbations addressed by $l_\infty$ adversarial training.

To validate this observation, we further analyze adversarial images against the $l_\infty$-AT Faster R-CNN using DWT. The results are shown in \cref{fig:motivation_at}.
The detector was adversarially trained with a $l_\infty$ perturbation bound of \( \epsilon=\frac{8}{255} \).
The results indicate that both AdvPatch and AdvTexture have a significantly larger amplitude than benign image patches when they adaptively attack the AT model. Moreover, unlike the findings in \cref{fig:fsd_motivation}, where AdvPatch has an amplitude similar to benign images at frequency level $D_6$, the DWT amplitude of AdvPatch against the AT detector is significantly larger than benign image patches at all frequency levels. This suggests that a physically realizable attack has to bypass AT models with higher amplitude, allowing the DWT Preprocessing module to more effectively localize and blur out the adversarial regions.

Therefore, we use the popular $l_\infty$-AT model to enhance defenses against low-amplitude adversarial signals.In our framework, the ASD masking module detects and suppresses high-amplitude adversarial signals in the frequency domain, while adversarial training improves robustness against low-amplitude perturbations in the image space. Together, these two components address adversarial perturbations across different magnitudes and spatial frequencies.

Importantly, ASD and AT operate at different stages of the pipeline and can be integrated without re-training the model. ASD performs input preprocessing, while AT improves the robustness of the detector itself. Therefore, we directly adopt the off-the-shelf $l_\infty$ adversarially trained Faster R-CNN from \citet{li2023od-at} in our experiments.

The complementary roles of ASD and AT in defending against perturbations with different magnitudes and spatial frequencies are illustrated in \cref{fig:illust_full_spectrum}. ASD with smaller $n$ deals with high-frequency perturbation, and needs high amplitude to identify. ASD with larger $n$ similarly deals with low-frequency high-amplitude perturbation. ASD is grounded with off-the-shelf AT model to defend against low-amplitude perturbation regardless of frequency.

\section{Experiments}

In this section, we show the defense performance of ASD with extensive experiments against both patch-based and texture-based attacks.\footnote{Code available at: \url{https://github.com/weiz0823/adv-spectral-defense}.}

\subsection{Experimental Setup}
We conducted experiments to evaluate the performance of object detection models when equipped with the defense method against a variety of physically realizable adversarial attacks.
The main detection models evaluated were Faster R-CNN \citep{frcnn}, FCOS \citep{fcos} and YOLOv8 \citep{yolov8}. Faster R-CNN represents a typical two-stage detector. FCOS and YOLOv8 represent two typical single-stage detectors. For standardly trained models, we used the models pretrained on MS-COCO \citep{coco}. For AT models, we used the checkpoint provided by \citet{li2023od-at}, which was the SOTA detector trained on MS-COCO with AT in the $l_\infty$ perturbation bound.
Since \citet{li2023od-at} did not provide AT checkpoints on YOLOv8, we reproduced their AT recipe on YOLOv8.
The original CSPDarknet-53 backbone had a severe trade-off between clean accuracy and robustness, so we used the ResNet-50 backbone, consistent with other models in \citet{li2023od-at}.

\subsubsection{Evaluated attacks}
The attacks were divided into patch-based adversarial attacks \citep{thys2019fooling,xu2020adv-tshirt} and texture-based adversarial attacks \citep{hu2022adv-texture,hu2023camou}. For patch-based attacks, we evaluated by sticking the patches onto images \citep{thys2019fooling,xu2020adv-tshirt} in the Inria Person \citep{inria} dataset.
Note that both AdvPatch \citep{thys2019fooling} and AdvTshirt \citep{xu2020adv-tshirt} did not have any constraint on the patch texture.
They differ by the way of application of the patch onto person's body: AdvPatch targeted cardboard patch; AdvTshirt targeted patch on the deformable t-shirt surface.
For texture-based attacks, we used the 3D rendering pipeline from \citet{hu2023camou} to craft the attack, which rendered person models to perform the attack.
There is no unconstrained texture-based attack, so we selected two different constraints: AdvTexture \citep{hu2022adv-texture} without any texture constraints, but needs to be tileable; AdvCaT \citep{hu2023camou} with camouflage color and pattern constraint.
All input images were padded to equal height and width and were resized to $416\times 416$.
Visualization of the evaluated adversarial attacks are provided in \cref{fig:vis_examples}. All attacks need to generate universal perturbation that applies to all images and random transformations.
AdvPatch and AdvTshirt constrain the perturbation to rectangle area, so they need strong patterns to weigh out normal features, causing high amplitudes on spectrum.
AdvTexture and AdvCaT apply perturbation to whole clothing area. AdvCaT enforces the colors to be the four camouflage colors, enhancing stealthiness, and limiting amplitude on spectrum.
The threat model is explicitly defined as follows:
\begin{itemize}
\item \emph{Attacker's knowledge:}
We use white-box (adaptive) attacks to evaluate the defense methods.
The attacker has access to the full model along with the parameters, including the defense module.

\item \emph{Attacker's capabilities:}
The attacker can generate a universal patch or texture $\tilde{p}$, which is applied to a specific region of the image (\eg, body of the person) by a predefined transformation function $A$. The adversarial image is $\tilde{x}=x+A(\tilde{p})$.
\end{itemize}

In order to cover a broad threat landscape, we also add two imperceptible attacks, NumbOD \citep{zhou2025numbod} and LGP \citep{li2024lgp}. Note that neither attack is physically realizable. They add human imperceptible perturbation by limiting magnitude on pixels, and the perturbation is different for each image. They are the attacks that $l_\infty$-AT defends against.

\subsubsection{Baseline methods for comparison}
We implemented several SOTA defense methods as the compared baselines, as discussed in \cref{sec:related}. We mainly utilized official code of the defense methods as well as the hyperparameters, and tested them against the attacks in this paper. Note that although the implementation is the same as the original paper, we evaluated the different defenses against the same attacks for a fair comparison. The attacks may be different from the attacks used in the defense papers.

Among the baselines, LGS \citep{lgs-naseer2019}, SAC \citep{sac-liu2022}, Jedi \citep{jedi-tarchoun2023} and NAPGuard \citep{wu2024napguard} were preprocessing-based defense methods, which localized and masked out adversarial patches by different criteria.
APE \citep{ape-kim2022} was an in-processing defense method, which suppressed the adversarial effect by clipping the values of inner feature maps. FD-JPEG \citep{liu2019fd-jpeg} was compared as a baseline of frequency-based adversarial defense. Note that it was originally designed for and evaluated on digital adversarial attacks such as PGD \citep{madry2017pgd}.

To test those methods, we used the official code by the original papers where possible.
We used the original code and pretrained checkpoints from the repositories of SAC\footnote{https://github.com/joellliu/SegmentAndComplete} and NAPGuard\footnote{https://github.com/wsynuiag/NAPGaurd}. We used the official Matlab code of Jedi\footnote{https://github.com/ihsenLab/jedi-CVPR2023}.
We copied the core function \verb|FD_jpeg_encode| from the official implementation of FD-JPEG\footnote{https://github.com/zihaoliu123/Feature-Distillation-DNN-Oriented-JPEG-Compression-Against-Adversarial-Examples}.
For IDBD, APE, and LGS, we could not find official code, so we implemented them based on the descriptions in the original papers.
Since each paper evaluated different attack settings, we aligned them under the same attack setting in our experiments.

\subsubsection{Hyper-parameters of ASD}
We set maximum expansion level $n=3$, detection threshold $\theta=0.17$. We aggregated detection results from frequency levels $n=0,1,2,3$, and performed an NMS with an Intersection-over-Union (IoU) threshold of $0.4$ to aggregate those results.

\subsubsection{Evaluation metric}
Mean average precision (mAP) is the widely used metric to evaluate object detection models. We evaluated mAP with intersection-over-union (IoU) threshold of 0.5, denoted mAP\textsubscript{50}. Since we have only one target class of \textit{person}, mAP\textsubscript{50} is also denoted as AP\textsubscript{50} without ambiguity.

\subsection{Adaptive Attack Design}
\label{sec:expr_adaptive}
In order to reliably evaluate the performance of a defense method, adaptive attack, which takes white-box knowledge of the defense method into account, should be evaluated \citep{obfuscated,tramer2020adaptive,autoattack}.
We designed adaptive attacks mainly following the principles of \citet{obfuscated}.
Moreover, we evaluated three types of adaptive attacks specially designed against ASD.
The first type of adaptive attack was the direct adaptive attack, which took the gradient of the defended model to optimize the adversarial input.
The second type of adaptive attack targeted the adversarial region detection module in ASD.
We added a loss term $L_\text{DWT}$ to suppress the detection of the adversarial pattern.
The third type of adaptive attack limited the amplitude of DWT coefficients by limiting the dynamic range of adversarial pattern, thereby avoiding the detection by DWT decomposition.
In the results, we show the performance of ASD against the strongest adaptive attack (the lowest AP).

\subsubsection{Adaptive attack designs of baseline methods}
\label{sec:adaptive_baseline}
We mainly followed the design principles of adaptive attacks by \citet{obfuscated}. LGS, APE, AT were evaluated with direct adaptive attack that directly backpropagated on the defended model. FD-JPEG, IDBD, Jedi, SAC were adaptively attacked with the straight-through estimator \citep{obfuscated}. We used identity mapping to estimate the four preprocess-based defense methods.
NAPGuard used an extra detector to localize patch regions.
We adaptively attacked it with an extra loss term on the extra detector.

\subsubsection{Adaptive attack designs against ASD}
\label{sec:adaptive_design_asd}
Three types of adaptive attacks specially designed against ASD are implemented.

The first type of adaptive attack was the \emph{direct} adaptive attack, which took the gradient of the defended model to optimize the adversarial input. Since ASD preprocessed the input images with Gaussian blurring in specific regions, which was implemented as a convolution with blurring kernel, it preserved the gradient. Therefore, we implemented this kind of adaptive attack as backpropagating through both the detector and the preprocessor.

The second type of adaptive attack targeted the adversarial region detection module in ASD. We tried to suppress the detection of the adversarial pattern by adding a loss term to reduce the coefficients of the DWT feature map. We denote this kind of adaptive attack as \emph{loss}. The loss term $L_\text{DWT}$ is formulated as
\begin{equation}
    L_\text{DWT}=\sum_{i=1}^n \frac{1}{2^i}(\norm{D_i^h}_2^2+\norm{D_i^v}_2^2+\norm{D_i^d}_2^2).
\end{equation}
The overall loss is
\begin{equation}
    L=L_\text{orig}+\lambda_\text{DWT} L_\text{DWT},
\end{equation}
where $L_\text{orig}$ is the original loss to craft the attack, and the coefficient was set to $\lambda_\text{DWT}=1$.

Since ASD only detects regions with high DWT amplitude, we limit the dynamic range of adversarial pattern in the third type of adaptive attack. We denote this kind of adaptive attack as \emph{range}.
Given white-box knowledge of defense hyper-parameter $\theta$, the maximum dynamic range for the adversarial pattern not to be detected is $\theta$. The pixel value range was set to $[0.5-\flatfrac{\theta}{2}, 0.5+\flatfrac{\theta}{2}]$.

\begin{table}[!t]
    \centering
    \caption{AP\textsubscript{50} (\%) of Faster R-CNN against different types of adaptive attacks designed for ASD.}
    \begin{tabular}{ccc}
        \toprule
        Defense & Adaptive type & AdvPatch \\
        \midrule
        \multirow{3}{*}{ASD} & Direct & 71.68 \\
        & Loss & 71.85 \\
        & Range & 58.78 \\
        \midrule
        \multirow{3}{*}{ASD + AT} & Direct & 90.04 \\
        & Loss & 88.00 \\
        & Range & 92.71 \\
        \bottomrule
    \end{tabular}
    \label{tab:adaptive_2}
\end{table}

We evaluated ASD against all three types of adaptive attack in the typical AdvPatch setting.
The results are shown in \cref{tab:adaptive_2}. In the worst case, standardly trained model with ASD got only an AP of \SI{58.78}{\percent}.
This was the result of adaptive attack with limited input range.
The results show that against the standardly trained model, the limited data range crafted adversarial pattern that had low DWT amplitude (lower than benign image patches), however this was also be adversarial to the standardly trained model.
ASD+AT achieved high APs against all three adaptive attacks specifically designed for ASD.
The most effective adaptive attack was the one with $L_\text{DWT}$, where the AP was \SI{88.00}{\percent}.

\subsection{Defense Performance in the Digital World}
\label{sec:result_main}
\begin{table*}[!t]
    \centering
    \caption{AP\textsubscript{50} (\%) of adversarially trained Faster R-CNN \citep{frcnn} against physically realizable (patch, texture) and imperceptible adversarial attacks in adaptive attack settings.}
    \begin{tabular}{ccccccccc}
        \toprule
        & & \multicolumn{2}{c}{Patch} & \multicolumn{2}{c}{Texture} & \multicolumn{2}{c}{Imperceptible} \\
        \cmidrule(lr){3-4}\cmidrule(lr){5-6}\cmidrule(lr){7-8}
        Defense & Clean & AdvPatch \citep{thys2019fooling} & AdvTshirt \citep{xu2020adv-tshirt} & AdvTexture \citep{hu2022adv-texture} & AdvCaT \citep{hu2023camou} & NumbOD \citep{zhou2025numbod} & LGP \citep{li2024lgp} & Average \\
        \midrule
        $l_\infty$-AT \citep{li2023od-at} & 95.96 & 55.46 & 9.59 & 29.08 & 39.59 & 93.31 & 94.38 & 53.57 \\
        \midrule
        FD-JPEG \citep{liu2019fd-jpeg} + AT & 89.92 & 52.59 & 8.04 & 25.98 & 39.17 & 84.44 & 84.22 & 49.07 \\
        IDBD \citep{epgf-zhou2020} + AT & 95.28 & 48.24 & 8.48 & 35.80 & 32.55 & 92.25 & 93.07 & 51.73 \\
        SAC \citep{sac-liu2022} + AT & 95.96 & 55.46 & 9.01 & 26.05 & 40.98 & 93.21 & 94.50 & 53.20 \\
        Jedi \citep{jedi-tarchoun2023} + AT & 91.16 & 72.04 & 30.17 & 28.59 & 40.65 & 91.62 & 92.40 & 59.25 \\
        APE \citep{ape-kim2022} + AT & 95.99 & 58.28 & 41.96 & 35.64 & 41.98 & $\mathbf{94.01}$ & 94.09 & 60.99 \\
        LGS \citep{lgs-naseer2019} + AT & 95.04 & 63.27 & 35.41 & 44.63 & 44.10 & 93.09 & 93.41 & 62.32 \\
        NAPGuard \citep{wu2024napguard} + AT & 96.07 & 82.13 & 36.84 & 27.57 & 41.31 & 93.52 & 94.47 & 62.64  \\
        \midrule
        ASD + AT & $\mathbf{96.12}$ & $\mathbf{88.00}$ & $\mathbf{62.04}$ & $\mathbf{70.08}$ & $\mathbf{54.64}$ & 93.23 & $\mathbf{94.66}$ & $\mathbf{77.11}$ \\
        \bottomrule
    \end{tabular}
    \label{tab:diverse_attacks}
\end{table*}

Usually, turning adversarial patterns crafted in the digital world to the physical world would degrade the attack ability due to complex physical factors including manufacturing process, light, distance, and so on. Thus, defending attacks in the digital world is harder than that in the physical world.

We evaluated the detection performance of the adversarially trained Faster R-CNN when equipped with an additional defense method, and the results are shown in \cref{tab:diverse_attacks}.
\emph{Clean} denotes the APs on the Inria \citep{inria} dataset, without any patch applied. Patch attacks were evaluated on the Inria dataset. Texture attacks were evaluated on the rendered images with adversarial clothing textures, following the settings of \citet{hu2023camou}.
\emph{Average} denotes the average AP against all six attacks.
All baselines and ASD were evaluated using adaptive attacks.
On the clean Inria dataset, ASD+AT did not lower the performance of Faster R-CNN, and obtained an AP of \SI{96.12}{\percent}.
We also checked that the mAP\textsubscript{50:95}, i.e. mAP averaged over IoU thresholds from 0.5 to 0.95, on COCO \citep{coco} dataset only lowered from 0.374 to 0.349.
Against the four attacks that covered both patch and texture-based attacks, the detector with ASD+AT obtained an average AP of \SI{68.69}{\percent}, outperforming previous defense methods by \SI{21.73}{\percent}.
Against all six attacks -- four physically realizable and two imperceptible -- the detector with ASD+AT obtained an average AP of \SI{77.11}{\percent}, outperforming previous defense methods by \SI{14.47}{\percent}.
Specifically, against AdvTshirt, whose adversarial patch covered a larger area than AdvPatch, the AP of ASD+AT only dropped \SI{25.96}{\percent}, a drop much lower than baseline methods.
It demonstrates that the performance of ASD+AT is generalizable to attack methods that cover a larger area, since the adversarial region localization of ASD is agnostic to patch size.
The results demonstrate that the proposed ASD framework generalizes effectively across diverse threat settings in the current threat landscape.
Below, we focus on physically realizable attacks, which is what ASD is designed for.

\paragraph*{More detectors}
We further tested the performance of the FCOS detector and the YOLOv8 detector with ASD+AT. The results of FCOS and YOLOv8 are shown in \cref{tab:diverse_attacks_fcos,tab:main_yolov8_at}.
The conclusions are similar to those of the Faster R-CNN detector. On the FCOS detector, ASD+AT obtained an average AP of \SI{56.23}{\percent}, outperforming previous defense methods by \SI{21.78}{\percent}. On the YOLOv8 detector, ASD+AT obtained an average AP of \SI{57.81}{\percent}, outperforming previous defense methods by \SI{10.15}{\percent}.

\begin{table*}[!t]
    \centering
    \caption{AP\textsubscript{50} (\%) of adversarially trained FCOS \citep{fcos} against physically realizable adversarial attacks in the adaptive attack settings. The same conventions are used as in \cref{tab:diverse_attacks}.}
    \begin{tabular}{ccccccc}
        \toprule
        & & \multicolumn{2}{c}{Patch} & \multicolumn{2}{c}{Texture} \\
        \cmidrule(lr){3-4}\cmidrule(lr){5-6}
        Defense & Clean & AdvPatch & AdvTshirt & AdvTexture & AdvCaT & Average \\
        \midrule
        $l_\infty$-AT & 93.87 & 30.54 & 7.17 & 30.07 & 18.33 & 21.53 \\
        \midrule
        IDBD + AT & 93.65 & 29.08 & 6.97 & 27.27 & 17.82 & 20.28 \\
        FD-JPEG + AT & 88.64 & 28.62 & 6.16 & 26.55 & 16.28 & 19.40 \\
        SAC + AT & 93.87 & 30.97 & 7.36 & 26.89 & 19.29 & 21.13 \\
        Jedi + AT & 90.35 & 37.20 & 18.16 & 21.52 & 9.48 & 21.59 \\
        APE + AT & 94.23 & 58.85 & 27.39 & 33.10 & 18.47 & 34.45 \\
        LGS + AT & 93.56 & 41.89 & 13.30 & 44.03 & 23.59 & 30.70 \\
        NAPGuard + AT & 94.54 & 64.62 & 10.34 & 28.00 & 18.57 & 30.38 \\
        \midrule
        ASD + AT & $\mathbf{94.79}$ & $\mathbf{77.52}$ & $\mathbf{49.02}$ & $\mathbf{66.60}$ & $\mathbf{31.77}$ & $\mathbf{56.23}$ \\
        \bottomrule
    \end{tabular}
    \label{tab:diverse_attacks_fcos}
\end{table*}
\begin{table*}[!t]
    \centering
    \caption{AP\textsubscript{50} (\%) of adversarially trained YOLOv8 \citep{yolov8} against patch-based and texture-based attacks in the adaptive attack settings.
    The same conventions are used as in \cref{tab:diverse_attacks}.}
    \begin{tabular}{ccccccc}
        \toprule
        & & \multicolumn{2}{c}{Patch} & \multicolumn{2}{c}{Texture} \\
        \cmidrule(lr){3-4}\cmidrule(lr){5-6}
        Defense & Clean & AdvPatch & AdvTshirt & AdvTexture & AdvCaT & Average \\
        \midrule
        $l_\infty$-AT & 66.25 & 38.32 & 43.50 & 57.39 & 51.42 & 47.66 \\
        \midrule
        IDBD + AT & 66.18 & 36.22 & 20.51 & 58.45 & 53.01 & 42.05 \\
        FD-JPEG + AT & 58.73 & 34.13 & 19.97 & 52.06 & 46.13 & 38.07 \\
        SAC + AT & $\mathbf{66.26}$ & 38.16 & 22.20 & 52.89 & 52.68 & 41.48 \\
        Jedi + AT & 49.27 & 43.67 & 25.89 & 55.06 & 55.38 & 45.00 \\
        APE + AT & 65.86 & 43.61 & 25.43 & 58.03 & 52.05 & 44.78 \\
        LGS + AT & 62.00 & 47.93 & 37.58 & 66.64 & $\mathbf{60.42}$ & 53.14 \\
        NAPGuard + AT & 65.05 & 48.82 & 25.86 & 56.13 & 53.34 & 46.04 \\
        \midrule
        ASD + AT & 64.70 & $\mathbf{60.49}$ & $\mathbf{50.38}$ & $\mathbf{67.29}$ & 53.07 & $\mathbf{57.81}$ \\
        \bottomrule
    \end{tabular}
    \label{tab:main_yolov8_at}
\end{table*}

\subsection{Defense Performance in the Physical World}
\begin{table}[!t]
    \centering
    \caption{AP\textsubscript{50} (\%) of Faster R-CNN \citep{frcnn} against texture-based attacks in the physical world.}
    \begin{tabular}{ccc}
        \toprule
        Defense & AdvTexture & AdvCaT \\
        \midrule
        None & 0.23 & 5.94 \\
        IDBD & 6.97 & 7.92 \\
        FD-JPEG & 0.19 & 18.81 \\
        SAC & 60.80 & 5.94 \\
        Jedi & 80.36 & 92.85 \\
        APE & 0.73 & 9.80 \\
        LGS & 0.44 & 98.83 \\
        NAPGuard & 12.96 & 4.62 \\
        \midrule
        ASD & 85.94 & 18.95 \\
        \midrule
        AT & 99.01 & 100.00 \\
        ASD + AT & $\mathbf{100.00}$ & $\mathbf{100.00}$ \\
        \bottomrule
    \end{tabular}
    \label{tab:result_physical}
\end{table}

One may wonder the performance of our ASD+AT facing the adversarial examples in the physical world. We evaluated different defense methods on the recorded videos of the AdvTexture \citep{hu2022adv-texture} and AdvCaT \citep{hu2023camou} from the authors, which corresponds to the non-adaptive attack setting.
For each attack method, 6 videos were evaluated. The videos included 2 scenes: indoor and outdoor, and 3 actors for each scene.
32 frames were evenly extracted from each video, thus the viewing angles were evenly distributed. Thus, 192 images were evaluated for each attack. All images were padded to square shape and resized to $416 \times 416$. The physical-world adversarial clothes of both attacks were crafted against the undefended Faster R-CNN detectors.

All defense methods were applied with the same hyper-parameters as in the evaluation against digital-world attacks.
The results are provided in \cref{tab:result_physical}. Since the recorded videos represent the non-adaptive attack setting, which represents weaker but more realistic setting compared to the adaptive attack setting used in the digital world,
AT alone obtained very high APs, and ASD+AT achieved both \SI{100}{\percent} defense performance against both attacks.
The results demonstrate that the defense performance also generalizes to the physical world.

\subsection{Extended Evaluation in the Digital World}
We conducted extended evaluation under various settings in the digital world to demonstrate the generalization of defense performance.

\subsubsection{Defense performance against non-adaptive attacks}
Besides evaluating the worst-case robustness against the strongest adaptive attack,
here we also consider a weaker but more realistic adversary -- non-adaptive attack.
Non-adaptive attack means that the attacker does not know the defense applied to the detector.
Thus, this setting is evaluated by transferring the adversarial examples crafted on the undefended detector to the detector with defense applied.
Against AdvPatch, ASD achieved an AP of \SI{83.75}{\percent}, and ASD + AT achieved an AP of \SI{93.21}{\percent}. Against AdvTexture, ASD and ASD + AT achieved APs of \SI{85.94}{\percent} and \SI{100.00}{\percent}, respectively. The results demonstrate that ASD also had a good performance against non-adaptive attacks.

\begin{table}[!t]
    \centering
    \caption{AP\textsubscript{50} (\%) of Faster R-CNN with ASD+AT against different NatPatch patches. Patches are denoted with the names in the original paper \citep{hu2021nat-patch}.}
    \begin{tabular}{cccccc}
        \toprule
        Patch & V2 & V3 & V4 & V4t & FR \\
        \midrule
        AP\textsubscript{50} (\%) & 92.67 & 92.07 & 89.62 & 90.33 & 90.56 \\
        \bottomrule
    \end{tabular}
    \label{tab:nat_patch}
\end{table}
\subsubsection{Defense performance against naturalistic patches}
Despite having evaluated the performance against unconstrained patches, we also included a type of patch that have constraint on the naturalness of the patch, denoted NatPatch \citep{hu2021nat-patch}.
We evaluated the performance against different pretrained natural patches in NatPatch, and the results are shown in \cref{tab:nat_patch}. The results show that ASD+AT achieved APs of at least \SI{89.62}{\percent} against different natural patches, demonstrating that ASD+AT also had a good performance against natural patch attacks.

\subsection{Ablation Study}
\label{sec:expr_ablation}

\begin{table}[!t]
    \centering
    \caption{Ablation study on ASD, evaluated against AdvPatch attack. Target model is Faster R-CNN.}
    \begin{tabular}{cc}
        \toprule
        Configuration of ASD & AP\textsubscript{50} (\%) \\
        \midrule
        Full configuration & $\mathbf{88.00}$ \\
        Without AT & 58.78 {\scriptsize \textcolor{red}{($\downarrow$29.22)}} \\
        Without level aggregation ($k=1$) & 78.91 {\scriptsize \textcolor{red}{($\downarrow$9.09)}} \\
        Single-spectral masking ($n=1, k=1$) & 75.67 {\scriptsize \textcolor{red}{($\downarrow$12.33)}} \\
        \bottomrule
    \end{tabular}
    \label{tab:ablation_main}
\end{table}

\subsubsection{Effectiveness of different components}
To demonstrate the effectiveness of each component in the proposed ASD framework, we conducted ablation study, and the results are shown in \cref{tab:ablation_main}. Numbers in parentheses indicate performance change relative to the full configuration.
Without AT, the defense was evaded by adaptive attack that found low-magnitude adversarial pattern, causing a \SI{29.22}{\percent} drop in AP. Without level aggregation (\cref{sec:method_level_agg}), the AP decreased by \SI{9.09}{\percent}. With only single-spectral masking, i.e., single level of DWT expansion ($n=1$) instead of multiple levels (\cref{sec:fsd_method_single_level}), the AP dropped by \SI{12.33}{\percent}. This is because ASD relies on multiple levels of DWT expansion to capture adversarial patterns in different frequency bands.

\begin{table}[!t]
    \centering
    \caption{AP\textsubscript{50} (\%) of adversarially trained Faster R-CNN with ASD with different threshold $\theta$, evaluated against AdvPatch attack.}
    \begin{tabular}{ccccccc}
        \toprule
        $\theta$ & 0.05 & 0.10 & 0.15 & 0.17 & 0.20 & 0.30 \\
        \midrule
        AP\textsubscript{50} (\%) & 71.21 & 81.99 & 86.91 & $\textbf{88.00}$ & 87.71 & 80.91 \\
        \bottomrule
    \end{tabular}
    \label{tab:ablation_theta}
\end{table}

\subsubsection{Selection of threshold $\theta$}
We evaluated the performance of Faster R-CNN defended by ASD with the threshold $\theta$ set to different values. The results are shown in \cref{tab:ablation_theta}.
Either too large or too small value of $\theta$ results in degradation of the performance.
The performance is steady for $\theta\in [0.15, 0.2]$.
Therefore, we chose an appropriate threshold $\theta=0.17$ in our defense design.
All datasets have images normalized to $[0, 1]$, which means that the DWT magnitudes have similar distribution, so the value does not have to be tuned when the threat or the detector changes.

\begin{table}[!t]
    \centering
    \caption{AP\textsubscript{50} (\%) of adversarially trained Faster R-CNN with ASD with different expansion levels $n$, evaluated against AdvPatch attack.}
    \begin{tabular}{cccccc}
        \toprule
        $n$ & 1 & 2 & 3 & 4 & 5 \\
        \midrule
        Without level aggregation & 75.67 & 81.46 & 78.91 & 67.30 & 58.99 \\
        With level aggregation & 74.12 & 83.64 & 88.00 & 88.52 & 88.34 \\
        \bottomrule
    \end{tabular}
    \label{tab:ablation_n}
\end{table}

\subsubsection{Ablation study on expansion level $n$}
We conducted ablation study on the expansion level $n$, and the results are shown in \cref{tab:ablation_n}. Larger $n$ means more level of details, and the spatial frequency of the inspected features is lower. The first row shows the result without level aggregation. $k$ is set to 1 and only details on level $n$ is used. The defense performance was best with $n$ neither too large nor too small, to capture proper frequency band with identifiable adversarial regions and not destroy benign regions.

\subsubsection{Ablation study on frequency level aggregation}
The second row of \cref{tab:ablation_n} shows the result with level aggregation. Here we set $k=n+1$ to include all levels of details from 0 to $n$. The performance increased as more levels are integrated, and saturated at $n=3$. Therefore, we selected $n=3$ in our defense design.

\begin{table}[!t]
    \centering
    \caption{AP\textsubscript{50} (\%) of adversarially trained Faster R-CNN with ASD and different time-frequency transformation methods, evaluated against AdvPatch attack.}
    \begin{tabular}{cccc}
        \toprule
        Transform & DWT & FFT & DCT \\
        \midrule
        AP\textsubscript{50} (\%) & 88.00 & 55.50 & 55.39 \\
        \bottomrule
    \end{tabular}
    \label{tab:ablation_fft}
\end{table}

\subsubsection{Ablation study on time-frequency transformation methods}
We conducted ablation study on time-frequency transformation methods used in the attack design. Besides DWT we used, we evaluated Fast Fourier Transform (FFT) and Discrete Cosine Transform (DCT). As shown in \cref{tab:ablation_fft}, FFT and DCT had similarly sub-optimal performance. The main difference is that FFT and DCT are global transformation methods. They can be used to analyze frequency-domain features but do not capture spatial information. Meanwhile, DWT captures both frequency characteristics and spatial information, and is the best fit of our defense.

\begin{table}[!t]
    \centering
    \caption{AP\textsubscript{50} (\%) of differently trained Faster R-CNN with ASD against attacks.}
    \begin{tabular}{cccc}
        \toprule
        Defense & AdvPatch & AdvTexture & NumbOD \\
        \midrule
        None & 36.99 & 0.21 & 12.25 \\
        ASD & 58.78 & 15.86 & 18.49 \\
        AT & 55.46 & 29.08 & $\textbf{93.31}$ \\
        ASD+AT & $\textbf{88.00}$ & $\textbf{70.08}$ & 93.23 \\
        \bottomrule
    \end{tabular}
    \label{tab:ablation_at}
\end{table}

\begin{table*}[!t]
    \centering
    \caption{AP\textsubscript{50} (\%) of standardly trained Faster R-CNN against patch-based and texture-based attacks in the adaptive attack settings.
    The same conventions are used as in \cref{tab:diverse_attacks}.}
    \begin{tabular}{ccccccc}
        \toprule
        & & \multicolumn{2}{c}{Patch} & \multicolumn{2}{c}{Texture} \\
        \cmidrule(lr){3-4}\cmidrule(lr){5-6}
        Defense & Clean & AdvPatch & AdvTshirt & AdvTexture & AdvCaT & Average \\
        \midrule
        None & 96.21 & 36.99 & 8.93 & 0.21 & 0.32 & 11.61 \\
        \midrule
        IDBD & 95.09 & 42.99 & 8.28 & 2.13 & 0.34 & 13.44 \\
        FD-JPEG & 90.34 & 30.84 & 6.66 & 0.15 & 0.38 & 9.51 \\
        SAC & 96.21 & 57.72 & 35.53 & 0.25 & 0.67 & 23.54 \\
        Jedi & 92.67 & $\mathbf{64.40}$ & 27.36 & 2.30 & 0.70 & 23.69 \\
        APE & 96.17 & 47.96 & 25.38 & 0.80 & 0.26 & 18.60 \\
        LGS & 95.93 & 24.76 & 3.51 & 3.91 & 4.25 & 9.11 \\
        NAPGuard & 96.10 & 47.00 & 10.86 & 2.17 & 0.43 & 15.12 \\
        \midrule
        ASD & $\mathbf{96.34}$ & 58.78 & $\mathbf{60.16}$ & $\mathbf{15.86}$ & $\mathbf{4.71}$ & $\mathbf{34.88}$ \\
        \bottomrule
    \end{tabular}
    \label{tab:main_frcnn_st}
\end{table*}

\subsubsection{Effectiveness of AT}
We conducted extended evaluation on ASD with or without AT.
\Cref{tab:ablation_at} shows the comparison of ASD with or without AT across diverse attack settings. We selected one attack from each category. ASD on its own increased APs by \SI{21.79}{\percent}, \SI{15.65}{\percent} and \SI{6.24}{\percent}, respectively. It demonstrates that ASD succeeded in defending against high-magnitude adversarial patterns.
AT further boosted the performance of ASD, demonstrating that AT complemented ASD and defended against low-magnitude adversarial patterns.

We also provide a full evaluation of ASD without AT along with baseline defense methods. The results are provided in \cref{tab:main_frcnn_st}.
ASD still outperformed previous defense methods on the standardly trained model, but the overall defense performance against texture-based attacks was less satisfactory, with all APs of only \SI{15.86}{\percent} against AdvTexture and \SI{4.71}{\percent} against AdvCaT.
The results verifies our conjecture that the attacks could evade ASD by utilizing perturbations with small DWT amplitudes, and provides a rationale to integrate ASD with AT model, as detailed in \cref{sec:reinforce_at}.

\begin{table}[!t]
    \centering
    \caption{Accuracy (\%) of ResNet-50 \citep{resnet} with defenses against Adversarial Patch \citep{adversarial-patch} targeting different ImageNet \citep{imagenet} classes.}
    \begin{tabular}{cccccc}
        \toprule
        Defense & Tench & Balloon & Umbrella & Throne & Teddy bear \\
        \midrule
        AT & 56.22 & 32.84 & 31.54 & 75.80 & 16.30 \\
        AT+ASD & 99.42 & 78.74 & 87.50 & 99.86 & 99.96 \\
        \bottomrule
    \end{tabular}
    \label{tab:ablation_classes}
\end{table}

\subsubsection{Different classes}
In our main experiments, we evaluated ASD against patch, texture, and imperceptible attacks designed to hide the \textit{person} class from the detector. A natural question is whether ASD generalizes to attacks targeting other classes.
To investigate this, we implemented another physically realizable attack, Adversarial Patch \citep{adversarial-patch}, targeting different classes. We used models pretrained on ImageNet-1k \citep{imagenet} and performed targeted attacks for multiple target classes. The patch occupied \SI{10}{\percent} of the image pixels and was optimized on 5,000 randomly sampled images from the ImageNet-1k validation set for 10 epochs. During evaluation, the patch was applied with random positions and random transformations (e.g., brightness, contrast, rotation), independently sampled from those used during training.
The accuracies under targeted attack, i.e. the portion of unsuccessful attack, are reported in \cref{tab:ablation_classes}, where each column corresponds to a target class. AT+ASD consistently outperforms AT across all classes by a clear margin, demonstrating that ASD generalizes well to attacks targeting different classes.

\subsubsection{DINO-based detector}
We further extended our evaluation to SOTA DINO-based detector -- RF-DETR \citep{rf-detr}. We used the same setting as in \cref{tab:main_frcnn_st} and evaluated AP\textsubscript{50} of RF-DETR. Without any defense, RF-DETR had an AP of \SI{72.74}{\percent} against AdvPatch attack. With ASD, RF-DETR had an AP of \SI{89.24}{\percent} against the same attack. The results indicate that ASD generalizes to modern DINO-based detector.

\subsection{Inference Time with ASD}
\label{sec:inference_time}
\begin{table}[!t]
    \centering
    \caption{Inference latency per image of Faster R-CNN with different frequency-based defense methods applied.}
    \begin{tabular}{ccc}
        \toprule
        Defense & Latency (\si{ms/im}) & AP\textsubscript{50} (\%) \\
        \midrule
        None & 22.5 & 55.46 \\
        FD-JPEG & 28.2 & 52.59 \\
        ASD (without level agg.) & 23.3 & 78.91 \\
        ASD (full) & 56.8 & 88.00 \\
        \bottomrule
    \end{tabular}
    \label{tab:infer_time}
\end{table}
For security-critical applications, it is inevitable that extra computation is consumed for better security.
We compared the inference time of Faster R-CNN with different frequency-based defense methods: FD-JPEG \citep{liu2019fd-jpeg} and ASD. For FD-JPEG, we used the official source code released by the authors. For ASD, we measured inference time with the full version and the single-level version, \ie, without frequency level aggregation. The experiments were done on a single Nvidia GeForce RTX 3080. The results are shown in \cref{tab:infer_time}. AP\textsubscript{50} is evaluated against AdvPatch attack. Without level aggregation, the Faster R-CNN with ASD introduced a \SI{3.6}{\percent} slowdown for an AP increase of +\SI{23.45}{\percent}. In comparison, FD-JPEG was \SI{25.3}{\percent} slower. This was because DWT with Haar wavelet had a simple form and was easy to parallelize. When four levels were aggregated, ASD traded a $2.5\times$ slowdown for a further AP boost of +\SI{9.09}{\percent}. This could be further accelerated with more parallel compute units in the inference hardware, because preprocessed inputs from multiple levels could be stacked and forwarded in one pass.

\section{Conclusion}
\label{sec:conclusion}
We propose a novel defense strategy, Adversarial Spectrum Defense (ASD), to defend against patch-based and texture-based adversarial attacks. ASD uses the tool of DWT to decompose the input image into various frequency components and incorporates all of them to detect and localize adversarial patterns of the attacks. When combined with the off-the-shelf AT model, ASD forms a comprehensive defense strategy. Experiments demonstrate that ASD+AT effectively defends against both patch-based and texture-based attacks, achieving SOTA performance, even in the face of strong adaptive adversaries specifically designed against ASD.

\appendix[A Proof of the Link between Perturbation Amplitude and DWT Amplitude]
\label{sec:proof_amplitude}

We hereby prove that when the perturbation amplitude is bounded by $\norm{\delta}_\infty\le \epsilon$, the DWT amplitude of the perturbation is also bounded by $\norm{D_{i}}_\infty\le 2^{i}\epsilon$, for all $i\ge1$.
This statement provides a link between perturbation amplitude in the input space and the DWT amplitude in the frequency components.

We denote the perturbation $\delta=X^*-X$, where $X^*$ is the adversarial image and $X$ is the benign image. We assume the $l_\infty$ perturbation bound is $\epsilon$, \ie, $\norm{\delta}_\infty\le \epsilon$. We assume the 2D DWT is performed with Haar wavelet, where the filters are \( L = \bqty{\frac{\sqrt{2}}{2}, \frac{\sqrt{2}}{2}} \) and \( H = \bqty{-\frac{\sqrt{2}}{2}, \frac{\sqrt{2}}{2}} \).

We use the induction method to prove.
Assume $\norm{A_i}_\infty\le 2^i\epsilon$ and $\norm{D_i}_\infty\le 2^i\epsilon$ for some $i\ge 0$.
We take one $2\times 2$ block, 
$\begin{pmatrix}
    A_{i,0,0} & A_{i,0,1} \\
    A_{i,1,0} & A_{i,1,1} \\
\end{pmatrix}=\begin{pmatrix}
    a & b \\
    c & d \\
\end{pmatrix}$,
in one convolve-and-downsample process in 2D DWT. Following the formulation of 2D DWT, we have
\begin{align}
D^h_{i+1,0,0}&=\frac{1}{2}(b+d-a-c),\\
D^v_{i+1,0,0}&=\frac{1}{2}(c+d-a-b),\\
D^d_{i+1,0,0}&=\frac{1}{2}(a+d-b-c),\\
A_{i+1,0,0}&=\frac{1}{2}(a+b+c+d).
\end{align}
We rewrite the operation to get $D_{i+1}$, which takes the $l_2$ norm over three channels $D^h_{i+1},D^v_{i+1},D^d_{i+1}$, as
\begin{equation}
\label{eq:di_l2_norm}
    (D_{i+1,0,0})^2=(D^h_{i+1,0,0})^2+(D^v_{i+1,0,0})^2+(D^d_{i+1,0,0})^2.
\end{equation}
The $l_\infty$ norm bound gives independent constraints on the four values $a,b,c,d$, as $-2^i\epsilon \le a,b,c,d\le 2^i\epsilon$.
By taking the derivative of $(D_{i+1,0,0})^2$ over $a,b,c,d$, we get
\begin{equation}
    \abs{D_{i+1,0,0}}\le 2^{i+1}\epsilon.
\end{equation}
The equation is satisfied when $\begin{pmatrix}
    a & b \\
    c & d \\
\end{pmatrix}=\pm 2^i\epsilon\begin{pmatrix}
    -1 & 1 \\
    -1 & 1 \\
\end{pmatrix},
\pm 2^i\epsilon\begin{pmatrix}
    -1 & -1 \\
    1 & 1 \\
\end{pmatrix}$, or
$\pm 2^i\epsilon\begin{pmatrix}
    1 & -1 \\
    -1 & 1 \\
\end{pmatrix}$, which correspond to horizontal, vertical, and diagonal details, respectively.

We also have
\begin{equation}
    \abs{A_{i+1,0,0}}\le 2^{i+1}\epsilon.
\end{equation}
The equation is satisfied when $\begin{pmatrix}
    a & b \\
    c & d \\
\end{pmatrix}=\pm 2^i\epsilon\begin{pmatrix}
    1 & 1 \\
    1 & 1 \\
\end{pmatrix}$.
For all $2\times 2$ blocks, the convolve-and-downsample process is done in the same manner and independently.
Therefore, we have $\norm{A_{i+1}}_\infty\le 2^{i+1}\epsilon$ and $\norm{D_{i+1}}_\infty\le 2^{i+1}\epsilon$.

By assigning $A_0 \coloneq\delta$, we have the initial value for the induction.

Therefore, we prove that when the perturbation bound is $\norm{\delta}_\infty\le \epsilon$, the details have amplitudes $\norm{D_{i}}_\infty\le 2^{i}\epsilon$ for all $i\ge1$.

With the above conclusion, a model that is robust against $l_\infty$-norm bounded adversarial perturbations (\eg $l_\infty$-AT models) implies that the model is robust against perturbations with small DWT amplitudes.

{
  \small
  \bibliographystyle{IEEEtranN}
  \bibliography{main}
}

\end{document}